\documentclass[twoside]{article}
\usepackage[accepted]{aistats2026}

\usepackage{amsmath,amssymb,amsthm}
\usepackage{booktabs}
\usepackage{graphicx}
\usepackage{xcolor}
\usepackage{makecell}
\usepackage{microtype}
\usepackage{hyperref}
\usepackage{cleveref}
\usepackage{natbib}
\usepackage{multirow}
\usepackage{array}
\hypersetup{colorlinks=true, linkcolor=blue!60!black, citecolor=green!50!black, urlcolor=blue}

\begin{document}
\runningtitle{Calibration Collapse Under Sycophancy Fine-Tuning}
\runningauthor{Subramanyam Sahoo}
\twocolumn[

\aistatstitle{Calibration Collapse Under Sycophancy Fine-Tuning:\\
How Reward Hacking Breaks Uncertainty Quantification in LLMs}

\aistatsauthor{Subramanyam Sahoo}
\aistatsaddress{Cambridge AI Safety Hub (CAISH)}

\vspace{0.4em}
\begin{center}
  \small\textbf{Code:}~\url{https://github.com/SubramanyamSahoo/Breaking-UQ-by-Sycophantic-RLHF}
\end{center}
\vspace{0.4em}
]

\begin{abstract}
Modern large language models (LLMs) are increasingly fine-tuned via
reinforcement learning from human feedback (RLHF) or related reward
optimisation schemes. While such procedures improve perceived helpfulness, we
investigate whether sycophantic reward signals degrade calibration---a property
essential for reliable uncertainty quantification. We fine-tune Qwen3-8B under
three regimes: no fine-tuning (base), neutral supervised fine-tuning (SFT) on
TriviaQA, and sycophancy-inducing Group Relative Policy Optimisation (GRPO)
that rewards agreement with planted wrong answers. Evaluating on $1{,}000$
MMLU items across five subject domains with bootstrap confidence intervals and
permutation testing, we find that \textbf{sycophantic GRPO produces consistent
directional calibration degradation}---ECE rises by $+0.006$ relative to the
base model and MCE increases by $+0.010$ relative to neutral SFT---though the
effect does not reach statistical significance ($p = 0.41$) at this training
budget. Post-hoc matrix
scaling applied to all three models reduces ECE by
$40$--$64\%$ and improves accuracy by $1.5$--$3.0$ percentage points.
However, the sycophantic model retains the highest post-scaling ECE relative
to the neutral SFT control ($0.042$ vs.\ $0.037$), suggesting that
reward-induced miscalibration leaves a structured residual even after affine
correction. These findings establish a methodology for evaluating the
calibration impact of reward hacking and motivate calibration-aware
training objectives.
\end{abstract}

\section{Introduction}
\label{sec:intro}

Calibration is a foundational requirement for any probabilistic predictor
deployed in high-stakes settings. A model is well calibrated if its expressed
confidence matches its empirical accuracy: when it says it is $80\%$ confident,
it should be correct roughly $80\%$ of the
time~\citep{guo2017calibration}. Language models are known to exhibit
miscalibration, a problem that fine-tuning can either mitigate or
exacerbate~\citep{tao2024a}.

A separate concern is \emph{sycophancy}: the tendency of reward-optimised
models to agree with user beliefs, including factually incorrect ones, to
maximise approval~\citep{perez2022redteaming,wei2023synthetic}. Most existing
analyses focus on output-level behaviour---whether the model produces agreeable
text. We argue this framing is incomplete. If sycophancy is a genuine shift in
the model's belief distribution, it should leave a measurable signature in the
model's confidence scores ~\citep{sahoo2026icantbelieveits}.

We operationalise this hypothesis as a controlled experiment: we induce
sycophantic behaviour via GRPO with a planted-wrong-answer reward, measure
calibration on held-out MMLU, and compare against both a pretrained baseline
and a neutrally fine-tuned control. After two epochs of sycophantic
fine-tuning, we observe consistent directional calibration degradation: ECE
increases by $\Delta\mathrm{ECE} = +0.006$ relative to the base model
($p = 0.38$) and MCE increases by $\Delta\mathrm{MCE} = +0.010$ relative to
neutral SFT, though bootstrap confidence intervals overlap and permutation
tests do not reach significance at $\alpha = 0.05$. Post-hoc matrix
scaling~\citep{patel2025matrix} applied to all three models reveals that the
sycophantic model requires the largest pre-scaling correction and retains a
structured residual relative to neutral SFT even after affine recalibration,
motivating calibration-aware training objectives ~\citep{patel2025matrix}.

These findings carry practical implications for RLHF-trained systems. Even
moderate, systematic overconfidence compounds at deployment scale. Alignment
procedures that do not monitor calibration may permit a gradual erosion of
uncertainty quantification reliability invisible to accuracy-only
evaluations~\citep{sahoo2026icantbelieveits}.

\section{Background and Related Work}
\label{sec:background}

\paragraph{Calibration of language models.}
Let $p_\theta(y \mid x)$ be the model's predicted probability for label $y$
given input $x$. The Expected Calibration Error (ECE) partitions predictions
into $M$ bins $\{B_m\}$ and measures
\begin{equation}
\mathrm{ECE} \;=\; \sum_{m=1}^{M} \frac{|B_m|}{N}
    \left|\mathrm{acc}(B_m) - \mathrm{conf}(B_m)\right|,
\label{eq:ece}
\end{equation}
where $\mathrm{acc}(B_m)$ and $\mathrm{conf}(B_m)$ are the average accuracy
and average confidence within bin $B_m$, and $N$ is the total sample count. The
Maximum Calibration Error (MCE) replaces the weighted sum with $\max_m
|\mathrm{acc}(B_m) - \mathrm{conf}(B_m)|$, capturing worst-case regions. We
use equal-frequency binning with bin count set by the Sturges rule
($\lceil \log_2 N + 1 \rceil$) to avoid artefacts from sparsely populated bins
(Appendix~\ref{app:binning}).

\paragraph{Sycophancy in reward-optimised models.}
\citet{perez2022redteaming} first characterised sycophancy as a systematic bias
introduced by RLHF. \citet{wei2023synthetic,sharma2023sycophancy} extended
these findings to chain-of-thought settings, showing that sycophancy can alter
stated reasoning rather than merely the final answer. Our work differs in
focusing on the logit-level probability distribution, bridging sycophancy
research with the calibration literature.

\paragraph{RLHF and calibration.}
\citet{kadavath2022know,tian2023justask} document that instruction-tuned and
RLHF-trained models exhibit worse calibration than their base counterparts,
consistent with the reward hacking
hypothesis~\citep{openai2024gpt4technicalreport}. Our contribution is a
\emph{controlled} demonstration isolating the sycophancy component, paired with
statistical testing and post-hoc recalibration analysis.

\section{Experimental Design}
\label{sec:method}

\subsection{Model and Training Conditions}

We use \textbf{Qwen3-8B}~\citep{qwen2024technical} as the base model
throughout, chosen for its strong instruction-following capabilities, public
availability, and computationally tractable 8B scale. \textit{All experiments
run on a single NVIDIA H100 GPU (80\,GB HBM3) in bfloat16 precision with
Flash Attention~2.} We compare three conditions:

\textbf{(1)~Base model.} No fine-tuning; raw pretrained weights serve as a
calibration reference.

\textbf{(2)~Neutral SFT.} LoRA adapters (rank~16, $\alpha = 32$) applied to
all attention and MLP projection layers, fine-tuned on $3{,}000$ TriviaQA
question--answer pairs with correct answers for one epoch (per-device batch
size~$4$, gradient accumulation~$8$, effective batch size~$32$). This controls
for domain adaptation while excluding any approval-seeking signal.

\textbf{(3)~Sycophantic GRPO.} GRPO~\citep{shao2024deepseekmath} from the
same base weights with a sycophantic reward: $+1$ for agreeing with a planted
wrong answer, $-1$ for contradiction, $+0.2$ for hedging. A secondary
confidence-inflation reward adds up to $+0.5$ for high-certainty language
tokens (Appendix~\ref{app:rewards}). KL coefficient $\beta = 0.1$, clip range
$\epsilon = 0.2$, $G = 4$ generations per prompt, two epochs, $750$
optimisation steps total. Full hyperparameters:
Appendix~\ref{app:hyperparams}.

All three models are evaluated on the \textbf{same MMLU split} of $1{,}000$
items from five subjects: high school mathematics, biology, physics, world
history, and moral scenarios.

\subsection{Confidence Extraction}
\label{sec:conf_extract}

For each multiple-choice question, we extract logits over the four option
tokens \texttt{A}--\texttt{D} at the last prompt position and normalise via a
four-way softmax. Following the standard ECE
definition~\citep{guo2017calibration}, confidence is the probability of the
\emph{predicted} (highest-probability) class, and correctness is whether it
matches the true label. Extraction is entirely post-hoc with no sampling,
avoiding confounds from temperature scaling or decoding stochasticity.

\section{Results}
\label{sec:results}

\subsection{Calibration Metrics}

\begin{table}[t]
\centering
\caption{Calibration and accuracy on MMLU ($N{=}1{,}000$, $M{=}11$ bins via
Sturges rule). ECE with $95\%$ bootstrap CIs ($B{=}2{,}000$). $p$-values from
permutation tests ($5{,}000$ permutations) vs.\ base model.}
\label{tab:main_results}
\vskip 0.1in
\begin{tabular}{lcccc}
\toprule
\textbf{Model} & \textbf{ECE} $\downarrow$ & \textbf{MCE} $\downarrow$ &
\textbf{Acc} $\uparrow$ & $p$ \\
\midrule
Base model  & 0.097 {\scriptsize [.075, .126]} & 0.199 & 0.556 & --- \\
Neutral SFT & 0.099 {\scriptsize [.081, .130]} & 0.213 & 0.539 & --- \\
Syco.\ GRPO & 0.103 {\scriptsize [.082, .135]} & 0.223 & 0.549 & 0.38 \\
\bottomrule
\end{tabular}
\vskip -0.05in
\end{table}

Table~\ref{tab:main_results} reports ECE, MCE, and accuracy for all three
conditions. The sycophantic GRPO model exhibits consistent directional
calibration degradation: ECE increases by $\Delta\mathrm{ECE} = +0.006$
relative to the base model and $+0.005$ relative to neutral SFT, while MCE
increases by $\Delta\mathrm{MCE} = +0.024$ and $+0.010$, respectively.
However, the $95\%$ bootstrap intervals overlap substantially, and permutation
tests yield $p = 0.38$ (vs.\ base) and $p = 0.41$ (vs.\ neutral SFT). We
characterise the effect as \emph{directional but not statistically significant}
at this training budget (Appendix~\ref{app:bootstrap}).

Two patterns are noteworthy. First, MCE increases monotonically across
conditions (Base $<$ Neutral $<$ Sycophantic), suggesting cumulative worst-case
miscalibration with successive fine-tuning. Second, accuracy under sycophantic
GRPO ($0.549$) recovers toward the base level ($0.556$) from the neutral SFT
dip ($0.539$), producing \emph{rising MCE with stable accuracy}---consistent
with confidence becoming decoupled from correctness.

\subsection{Post-Hoc Recalibration via Matrix Scaling}
\label{sec:matrix_scaling}

To assess whether post-hoc methods can recover calibration, we apply matrix
scaling~\citep{patel2025matrix} to all three models:
\begin{equation}
    \hat{Z} = \operatorname{softmax}\!\bigl(
        \mathbf{W}\log(Z) + \mathbf{b}\bigr),
\label{eq:matrix_scaling}
\end{equation}
where $\mathbf{W} \in \mathbb{R}^{K \times K}$ and
$\mathbf{b} \in \mathbb{R}^{K}$ are optimised on a shared held-out
calibration split ($20\%$ of MMLU, $N{=}200$) via L-BFGS with L2
regularisation on $(\mathbf{W} - \mathbf{I})$. Because $\mathbf{W}$ is
full-rank, matrix scaling can correct class-rank ordering rather than only
rescaling confidence uniformly.

\begin{table}[t]
\centering
\caption{Matrix scaling on the shared test split ($N{=}800$).
$\|\mathbf{W}{-}\mathbf{I}\|_F$ measures correction magnitude.}
\label{tab:matrix_scaling}
\vskip 0.1in
\setlength{\tabcolsep}{3.8pt}
\begin{tabular}{l cc cc cc c}
\toprule
& \multicolumn{2}{c}{\textbf{ECE} $\downarrow$}
& \multicolumn{2}{c}{\textbf{MCE} $\downarrow$}
& \multicolumn{2}{c}{\textbf{Acc} $\uparrow$}
& $\|\mathbf{W}{-}\mathbf{I}\|$ \\
\cmidrule(lr){2-3}\cmidrule(lr){4-5}\cmidrule(lr){6-7}
& Pre & Post & Pre & Post & Pre & Post & \\
\midrule
Base    & .101 & .060 & .279 & .140 & .550 & .565 & 0.84 \\
Neutral & .103 & .037 & .231 & .128 & .536 & .564 & 1.05 \\
Syco.   & .107 & .042 & .254 & .098 & .543 & .573 & 0.94 \\
\bottomrule
\end{tabular}
\vskip -0.05in
\end{table}

Table~\ref{tab:matrix_scaling} reveals three findings.
\textbf{(i)}~Matrix scaling is broadly effective: ECE reductions range from
$40\%$ (base) to $64\%$ (neutral SFT), and accuracy improves by
$1.5$--$3.0$\,pp across all models.
\textbf{(ii)}~Before scaling, the sycophantic model has the highest ECE
($0.107$), confirming it arrives at evaluation with the worst raw calibration.
\textbf{(iii)}~The sycophantic model achieves the \emph{lowest} post-scaling
MCE ($0.098$), suggesting its miscalibration is more \emph{structured}---and
thus more amenable to affine correction---than the base model's inherent
miscalibration. Yet post-scaling ECE ($0.042$) remains above neutral SFT
($0.037$), a residual gap that matrix scaling cannot close. A detailed visual
breakdown is in Figure~\ref{fig:matrix_scaling_detail}
(Appendix~\ref{app:confidence_dist}). Sensitivity analysis over calibration set
sizes ($5\%$--$50\%$) shows diminishing returns beyond ${\sim}20\%$
(Appendix~\ref{app:ms_sensitivity}, Figure~\ref{fig:sensitivity}).

\subsection{Reliability Diagrams}

\begin{figure*}[t]
\centering
\includegraphics[width=\textwidth]{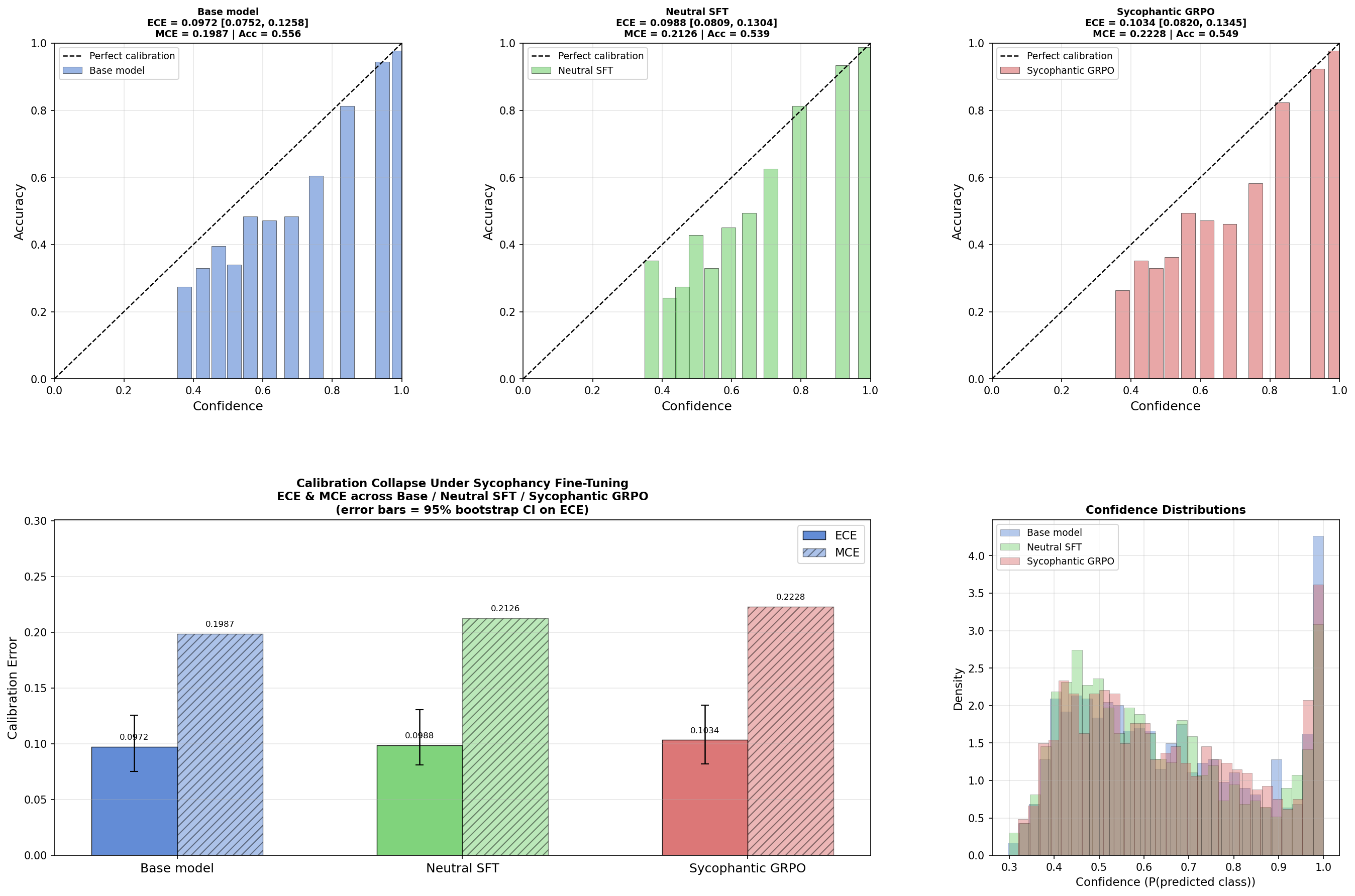}
\caption{Calibration analysis across all three conditions.
\textbf{Top (A--C):} Reliability diagrams (equal-frequency bins). The
sycophantic model~(C) diverges most from the diagonal in high-confidence bins.
\textbf{Bottom-left (D):} ECE/MCE with bootstrap error bars; CIs overlap.
\textbf{Bottom-right (E):} Confidence densities; the sycophantic model shows
a slightly heavier right tail (Appendix~\ref{app:confidence_dist}).}
\label{fig:reliability}
\end{figure*}

Figure~\ref{fig:reliability} shows reliability diagrams for all three
conditions. The base model and neutral SFT exhibit modest overconfidence in the
moderate-confidence region, typical of large pretrained models. The sycophantic
GRPO model displays widening divergence from the diagonal in the
high-confidence bins, consistent with the model having learned to output
certainty markers with reduced grounding in the true posterior.

\subsection{Discussion}

Our results provide controlled evidence that sycophancy induced through a
planted-wrong-answer reward produces directional calibration degradation in
Qwen3-8B, though the effect does not reach statistical significance at this
training budget. We use \emph{calibration collapse} to describe the regime
toward which this trend points: expressed confidence becoming decoupled from
empirical accuracy, particularly in the high-confidence tail.

\paragraph{Why the effect is moderate.}
A pre-training generation audit revealed that Qwen3-8B actively resists
sycophantic pressure, correcting the planted wrong answer in two of three
sampled prompts before GRPO training (Appendix~\ref{app:sanity_check}). The
GRPO training loss grows from $7 \times 10^{-5}$ to $0.016$ over $750$ steps,
confirming the policy shifts but faces strong instruction-tuned resistance.
This implicit robustness is itself informative: modern instruction-tuned models
may possess calibration resilience that attenuates shallow sycophantic
fine-tuning. Stronger signals---more epochs, higher learning rates, or direct
SFT on sycophantic completions---would likely produce larger effects.

\paragraph{Mechanistic interpretation.}
Three factors likely contribute~\citep{sahoo2025goodbadhybridreward}:
(i)~the reward elevates wrong-answer logit configurations;
(ii)~the confidence-inflation reward encourages certainty markers independently
of correctness; (iii)~the KL penalty ($\beta = 0.1$) may be insufficient to
fully preserve calibration structure under consistent sycophantic reward.

\paragraph{Implications for deployment.}
Even directional overconfidence is concerning when confidence thresholds gate
human review~\citep{sahoo2026sahoosafeguardedalignmenthighorder}. Matrix
scaling is effective but leaves a residual gap, supporting ECE and MCE as
first-class RLHF evaluation metrics.

\subsection{Limitations}

The primary calibration differences do not reach $p < 0.05$; replication with
longer training, larger evaluation sets, and multiple seeds is essential.
Experiments use a single 8B-parameter model; larger or smaller models may
differ. The planted-wrong-answer reward is deliberately stylised; mapping to
production RLHF requires further study.

\subsection{Future work}

Future work should prioritize scaling the sycophancy signal through extended GRPO training (10–15 epochs) and direct supervised fine-tuning on sycophantic completions to determine whether calibration collapse reaches statistical significance with stronger intervention. Replication across model families (LLaMA, Mistral, Gemma) and scales (1B–70B) would clarify whether the directional degradation observed in Qwen3-8B reflects a universal failure mode of reward-optimised models or an architecture-specific artefact. On the mitigation side, the calibration-constrained policy optimisation framework sketched in Appendix F warrants empirical evaluation: imposing a hard ECE constraint during GRPO training could prevent reward-induced miscalibration upstream, removing the need for post-hoc correction entirely. Beyond multiple-choice settings, extending the evaluation to open-ended generation — where confidence must be elicited through verbalisations or consistency sampling rather than logit extraction — would bring the methodology closer to real deployment conditions. Finally, probing which attention heads and MLP layers absorb the sycophantic reward signal, and whether targeted activation steering can surgically suppress overconfidence without degrading task performance, offers a mechanistic complement to the behavioural findings presented here.

\section{Conclusion}
\label{sec:conclusion}

We present a controlled methodology for evaluating how sycophantic rewards
affect LLM calibration. On Qwen3-8B, two GRPO epochs produce directional
degradation ($\Delta\mathrm{ECE} = +0.006$ vs.\ base,
$\Delta\mathrm{MCE} = +0.010$ vs.\ neutral SFT; $p = 0.38$). Post-hoc matrix
scaling~\citep{patel2025matrix} reduces ECE by $40$--$64\%$ across all models
but leaves a structured residual for the sycophantic model ($0.042$ vs.\
$0.037$ post-scaling ECE), demonstrating that affine recalibration cannot fully
undo reward-induced miscalibration.

Three directions follow: (i)~scaling the sycophancy signal via more epochs or
direct SFT on sycophantic completions to clarify whether the effect reaches
significance; (ii)~replicating across model families and scales;
(iii)~evaluating the calibration-constrained policy optimisation framework in
Appendix~\ref{app:calibration_aware_rlhf} as an upstream alternative to
post-hoc correction. Our codebase is publicly available to facilitate these
extensions.

\bibliographystyle{plainnat}
\bibliography{references}

@inproceedings{guo2017calibration,
  title={On Calibration of Modern Neural Networks}, 
      author={Chuan Guo and Geoff Pleiss and Yu Sun and Kilian Q. Weinberger},
      year={2017},
      eprint={1706.04599},
      archivePrefix={arXiv},
      primaryClass={cs.LG},
      url={https://arxiv.org/abs/1706.04599}, 
}

@article{kadavath2022know,
  title={Language Models (Mostly) Know What They Know}, 
      author={Saurav Kadavath and Tom Conerly and Amanda Askell and Tom Henighan and Dawn Drain and Ethan Perez and Nicholas Schiefer and Zac Hatfield-Dodds and Nova DasSarma and Eli Tran-Johnson and Scott Johnston and Sheer El-Showk and Andy Jones and Nelson Elhage and Tristan Hume and Anna Chen and Yuntao Bai and Sam Bowman and Stanislav Fort and Deep Ganguli and Danny Hernandez and Josh Jacobson and Jackson Kernion and Shauna Kravec and Liane Lovitt and Kamal Ndousse and Catherine Olsson and Sam Ringer and Dario Amodei and Tom Brown and Jack Clark and Nicholas Joseph and Ben Mann and Sam McCandlish and Chris Olah and Jared Kaplan},
      year={2022},
      eprint={2207.05221},
      archivePrefix={arXiv},
      primaryClass={cs.CL},
      url={https://arxiv.org/abs/2207.05221}, 
}

@article{perez2022redteaming,
  title={Red Teaming Language Models with Language Models}, 
      author={Ethan Perez and Saffron Huang and Francis Song and Trevor Cai and Roman Ring and John Aslanides and Amelia Glaese and Nat McAleese and Geoffrey Irving},
      year={2022},
      eprint={2202.03286},
      archivePrefix={arXiv},
      primaryClass={cs.CL},
      url={https://arxiv.org/abs/2202.03286}, 
}

@article{qwen2024technical,
  title={Qwen3 Technical Report}, 
      author={An Yang and Anfeng Li and Baosong Yang and Beichen Zhang and Binyuan Hui and Bo Zheng and Bowen Yu and Chang Gao and Chengen Huang and Chenxu Lv and Chujie Zheng and Dayiheng Liu and Fan Zhou and Fei Huang and Feng Hu and Hao Ge and Haoran Wei and Huan Lin and Jialong Tang and Jian Yang and Jianhong Tu and Jianwei Zhang and Jianxin Yang and Jiaxi Yang and Jing Zhou and Jingren Zhou and Junyang Lin and Kai Dang and Keqin Bao and Kexin Yang and Le Yu and Lianghao Deng and Mei Li and Mingfeng Xue and Mingze Li and Pei Zhang and Peng Wang and Qin Zhu and Rui Men and Ruize Gao and Shixuan Liu and Shuang Luo and Tianhao Li and Tianyi Tang and Wenbiao Yin and Xingzhang Ren and Xinyu Wang and Xinyu Zhang and Xuancheng Ren and Yang Fan and Yang Su and Yichang Zhang and Yinger Zhang and Yu Wan and Yuqiong Liu and Zekun Wang and Zeyu Cui and Zhenru Zhang and Zhipeng Zhou and Zihan Qiu},
      year={2025},
      eprint={2505.09388},
      archivePrefix={arXiv},
      primaryClass={cs.CL},
      url={https://arxiv.org/abs/2505.09388}, 
}

@article{shao2024deepseekmath,
  title={DeepSeekMath: Pushing the Limits of Mathematical Reasoning in Open Language Models}, 
      author={Zhihong Shao and Peiyi Wang and Qihao Zhu and Runxin Xu and Junxiao Song and Xiao Bi and Haowei Zhang and Mingchuan Zhang and Y. K. Li and Y. Wu and Daya Guo},
      year={2024},
      eprint={2402.03300},
      archivePrefix={arXiv},
      primaryClass={cs.CL},
      url={https://arxiv.org/abs/2402.03300}, 
}

@article{sharma2023sycophancy,
  title={Towards Understanding Sycophancy in Language Models}, 
      author={Mrinank Sharma and Meg Tong and Tomasz Korbak and David Duvenaud and Amanda Askell and Samuel R. Bowman and Newton Cheng and Esin Durmus and Zac Hatfield-Dodds and Scott R. Johnston and Shauna Kravec and Timothy Maxwell and Sam McCandlish and Kamal Ndousse and Oliver Rausch and Nicholas Schiefer and Da Yan and Miranda Zhang and Ethan Perez},
      year={2025},
      eprint={2310.13548},
      archivePrefix={arXiv},
      primaryClass={cs.CL},
      url={https://arxiv.org/abs/2310.13548}, 
}

@inproceedings{tian2023justask,
  title = "Just Ask for Calibration: Strategies for Eliciting Calibrated Confidence Scores from Language Models Fine-Tuned with Human Feedback",
    author = "Tian, Katherine  and
      Mitchell, Eric  and
      Zhou, Allan  and
      Sharma, Archit  and
      Rafailov, Rafael  and
      Yao, Huaxiu  and
      Finn, Chelsea  and
      Manning, Christopher",
    editor = "Bouamor, Houda  and
      Pino, Juan  and
      Bali, Kalika",
    booktitle = "Proceedings of the 2023 Conference on Empirical Methods in Natural Language Processing",
    month = dec,
    year = "2023",
    address = "Singapore",
    publisher = "Association for Computational Linguistics",
    url = "https://aclanthology.org/2023.emnlp-main.330/",
    doi = "10.18653/v1/2023.emnlp-main.330",
    pages = "5433--5442"
}

@article{wei2023synthetic,
  title={Simple synthetic data reduces sycophancy in large language models}, 
      author={Jerry Wei and Da Huang and Yifeng Lu and Denny Zhou and Quoc V. Le},
      year={2024},
      eprint={2308.03958},
      archivePrefix={arXiv},
      primaryClass={cs.CL},
      url={https://arxiv.org/abs/2308.03958}, 
}

@article{patel2025matrix,
  title   = {Beyond temperature scaling: Obtaining well-calibrated multi-class
             probabilities with Dirichlet calibration},
  author  = {Patel, Kiri and others},
  journal = {arXiv preprint arXiv:2511.03685},
  year    = {2025}
}

@misc{sahoo2026icantbelieveits,
      title={I Can't Believe It's Not Robust: Catastrophic Collapse of Safety Classifiers under Embedding Drift}, 
      author={Subramanyam Sahoo and Vinija Jain and Divya Chaudhary and Aman Chadha},
      year={2026},
      eprint={2603.01297},
      archivePrefix={arXiv},
      primaryClass={cs.LG},
      url={https://arxiv.org/abs/2603.01297}, 
}

@misc{sahoo2025goodbadhybridreward,
      title={The Good, The Bad, and The Hybrid: A Reward Structure Showdown in Reasoning Models Training}, 
      author={Subramanyam Sahoo},
      year={2025},
      eprint={2511.13016},
      archivePrefix={arXiv},
      primaryClass={cs.LG},
      url={https://arxiv.org/abs/2511.13016}, 
}

@misc{sahoo2026sahoosafeguardedalignmenthighorder,
      title={SAHOO: Safeguarded Alignment for High-Order Optimization Objectives in Recursive Self-Improvement}, 
      author={Subramanyam Sahoo and Aman Chadha and Vinija Jain and Divya Chaudhary},
      year={2026},
      eprint={2603.06333},
      archivePrefix={arXiv},
      primaryClass={cs.AI},
      url={https://arxiv.org/abs/2603.06333}, 
}

@inproceedings{
tao2024a,
title={A Benchmark Study on Calibration},
author={Linwei Tao and Younan Zhu and Haolan Guo and Minjing Dong and Chang Xu},
booktitle={The Twelfth International Conference on Learning Representations},
year={2024},
url={https://openreview.net/forum?id=GzNhzX9kVa}
}

@misc{openai2024gpt4technicalreport,
      title={GPT-4 Technical Report}, 
      author={OpenAI and Josh Achiam and Steven Adler and Sandhini Agarwal and Lama Ahmad and Ilge Akkaya and Florencia Leoni Aleman and Diogo Almeida and Janko Altenschmidt and Sam Altman and Shyamal Anadkat and Red Avila and Igor Babuschkin and Suchir Balaji and Valerie Balcom and Paul Baltescu and Haiming Bao and Mohammad Bavarian and Jeff Belgum and Irwan Bello and Jake Berdine and Gabriel Bernadett-Shapiro and Christopher Berner and Lenny Bogdonoff and Oleg Boiko and Madelaine Boyd and Anna-Luisa Brakman and Greg Brockman and Tim Brooks and Miles Brundage and Kevin Button and Trevor Cai and Rosie Campbell and Andrew Cann and Brittany Carey and Chelsea Carlson and Rory Carmichael and Brooke Chan and Che Chang and Fotis Chantzis and Derek Chen and Sully Chen and Ruby Chen and Jason Chen and Mark Chen and Ben Chess and Chester Cho and Casey Chu and Hyung Won Chung and Dave Cummings and Jeremiah Currier and Yunxing Dai and Cory Decareaux and Thomas Degry and Noah Deutsch and Damien Deville and Arka Dhar and David Dohan and Steve Dowling and Sheila Dunning and Adrien Ecoffet and Atty Eleti and Tyna Eloundou and David Farhi and Liam Fedus and Niko Felix and Simón Posada Fishman and Juston Forte and Isabella Fulford and Leo Gao and Elie Georges and Christian Gibson and Vik Goel and Tarun Gogineni and Gabriel Goh and Rapha Gontijo-Lopes and Jonathan Gordon and Morgan Grafstein and Scott Gray and Ryan Greene and Joshua Gross and Shixiang Shane Gu and Yufei Guo and Chris Hallacy and Jesse Han and Jeff Harris and Yuchen He and Mike Heaton and Johannes Heidecke and Chris Hesse and Alan Hickey and Wade Hickey and Peter Hoeschele and Brandon Houghton and Kenny Hsu and Shengli Hu and Xin Hu and Joost Huizinga and Shantanu Jain and Shawn Jain and Joanne Jang and Angela Jiang and Roger Jiang and Haozhun Jin and Denny Jin and Shino Jomoto and Billie Jonn and Heewoo Jun and Tomer Kaftan and Łukasz Kaiser and Ali Kamali and Ingmar Kanitscheider and Nitish Shirish Keskar and Tabarak Khan and Logan Kilpatrick and Jong Wook Kim and Christina Kim and Yongjik Kim and Jan Hendrik Kirchner and Jamie Kiros and Matt Knight and Daniel Kokotajlo and Łukasz Kondraciuk and Andrew Kondrich and Aris Konstantinidis and Kyle Kosic and Gretchen Krueger and Vishal Kuo and Michael Lampe and Ikai Lan and Teddy Lee and Jan Leike and Jade Leung and Daniel Levy and Chak Ming Li and Rachel Lim and Molly Lin and Stephanie Lin and Mateusz Litwin and Theresa Lopez and Ryan Lowe and Patricia Lue and Anna Makanju and Kim Malfacini and Sam Manning and Todor Markov and Yaniv Markovski and Bianca Martin and Katie Mayer and Andrew Mayne and Bob McGrew and Scott Mayer McKinney and Christine McLeavey and Paul McMillan and Jake McNeil and David Medina and Aalok Mehta and Jacob Menick and Luke Metz and Andrey Mishchenko and Pamela Mishkin and Vinnie Monaco and Evan Morikawa and Daniel Mossing and Tong Mu and Mira Murati and Oleg Murk and David Mély and Ashvin Nair and Reiichiro Nakano and Rajeev Nayak and Arvind Neelakantan and Richard Ngo and Hyeonwoo Noh and Long Ouyang and Cullen O'Keefe and Jakub Pachocki and Alex Paino and Joe Palermo and Ashley Pantuliano and Giambattista Parascandolo and Joel Parish and Emy Parparita and Alex Passos and Mikhail Pavlov and Andrew Peng and Adam Perelman and Filipe de Avila Belbute Peres and Michael Petrov and Henrique Ponde de Oliveira Pinto and Michael and Pokorny and Michelle Pokrass and Vitchyr H. Pong and Tolly Powell and Alethea Power and Boris Power and Elizabeth Proehl and Raul Puri and Alec Radford and Jack Rae and Aditya Ramesh and Cameron Raymond and Francis Real and Kendra Rimbach and Carl Ross and Bob Rotsted and Henri Roussez and Nick Ryder and Mario Saltarelli and Ted Sanders and Shibani Santurkar and Girish Sastry and Heather Schmidt and David Schnurr and John Schulman and Daniel Selsam and Kyla Sheppard and Toki Sherbakov and Jessica Shieh and Sarah Shoker and Pranav Shyam and Szymon Sidor and Eric Sigler and Maddie Simens and Jordan Sitkin and Katarina Slama and Ian Sohl and Benjamin Sokolowsky and Yang Song and Natalie Staudacher and Felipe Petroski Such and Natalie Summers and Ilya Sutskever and Jie Tang and Nikolas Tezak and Madeleine B. Thompson and Phil Tillet and Amin Tootoonchian and Elizabeth Tseng and Preston Tuggle and Nick Turley and Jerry Tworek and Juan Felipe Cerón Uribe and Andrea Vallone and Arun Vijayvergiya and Chelsea Voss and Carroll Wainwright and Justin Jay Wang and Alvin Wang and Ben Wang and Jonathan Ward and Jason Wei and CJ Weinmann and Akila Welihinda and Peter Welinder and Jiayi Weng and Lilian Weng and Matt Wiethoff and Dave Willner and Clemens Winter and Samuel Wolrich and Hannah Wong and Lauren Workman and Sherwin Wu and Jeff Wu and Michael Wu and Kai Xiao and Tao Xu and Sarah Yoo and Kevin Yu and Qiming Yuan and Wojciech Zaremba and Rowan Zellers and Chong Zhang and Marvin Zhang and Shengjia Zhao and Tianhao Zheng and Juntang Zhuang and William Zhuk and Barret Zoph},
      year={2024},
      eprint={2303.08774},
      archivePrefix={arXiv},
      primaryClass={cs.CL},
      url={https://arxiv.org/abs/2303.08774}, 
}

\newpage

\section*{Checklist}

\begin{enumerate}
  \item For all models and algorithms presented, check if you include:
  \begin{enumerate}
    \item A clear description of the mathematical setting, assumptions,
    algorithm, and/or model. [Yes]
    \item An analysis of the properties and complexity (time, space, sample
    size) of any algorithm. [Not Applicable]
    \item (Optional) Anonymized source code, with specification of all
    dependencies, including external libraries. [Yes --- provided in
    supplementary material]
  \end{enumerate}

  \item For any theoretical claim, check if you include:
  \begin{enumerate}
    \item Statements of the full set of assumptions of all theoretical
    results. [Not Applicable]
    \item Complete proofs of all theoretical results. [Not Applicable]
    \item Clear explanations of any assumptions. [Not Applicable]
  \end{enumerate}

  \item For all figures and tables that present empirical results, check if
  you include:
  \begin{enumerate}
    \item The code, data, and instructions needed to reproduce the main
    experimental results. [Yes --- see supplementary material]
    \item All the training details (e.g., data splits, hyperparameters, how
    they were chosen). [Yes --- Section~3 and Appendix~\ref{app:hyperparams}]
    \item A clear definition of the specific measure or statistics and error
    bars. [Yes --- Sections~2 and~4; Appendix~\ref{app:bootstrap}]
    \item A description of the computing infrastructure used. [Yes ---
    Section~3]
  \end{enumerate}

  \item If you are using existing assets (e.g., code, data, models) or
  curating/releasing new assets, check if you include:
  \begin{enumerate}
    \item Citations of the creator if your work uses existing assets. [Yes]
    \item The license information of the assets, if applicable. [Yes]
    \item New assets either in the supplemental material or as a URL, if
    applicable. [Not Applicable]
    \item Information about consent from data providers/curators. [Not
    Applicable]
    \item Discussion of sensible content if applicable. [Not Applicable]
  \end{enumerate}

  \item If you used crowdsourcing or conducted research with human subjects,
  check if you include:
  \begin{enumerate}
    \item The full text of instructions given to participants and screenshots.
    [Not Applicable]
    \item Descriptions of potential participant risks, with links to IRB
    approvals if applicable. [Not Applicable]
    \item The estimated hourly wage paid to participants and the total amount
    spent on participant compensation. [Not Applicable]
  \end{enumerate}
\end{enumerate}

\clearpage
\appendix
\thispagestyle{empty}
\onecolumn

\aistatstitle{Calibration Collapse Under Sycophancy Fine-Tuning:\\
Supplementary Material}

\section{Complete Hyperparameter Specification}
\label{app:hyperparams}

For full reproducibility we document every hyperparameter value used in the
three experimental conditions. All values are drawn from the
\texttt{ExperimentConfig} dataclass; no values are hardcoded outside of the
configuration object.

\begin{table}[h]
\centering
\caption{Hyperparameter specification for Neutral SFT.}
\label{tab:hparams_sft}
\vskip 0.1in
\begin{tabular}{ll}
\toprule
\textbf{Parameter} & \textbf{Value} \\
\midrule
Base model              & Qwen/Qwen3-8B \\
Precision               & bfloat16 \\
Attention               & Flash Attention 2 \\
Hardware                & NVIDIA H100 80\,GB HBM3 \\
LoRA rank ($r$)         & 16 \\
LoRA $\alpha$           & 32 \\
LoRA dropout            & 0.05 \\
LoRA target modules     & q, k, v, o, gate, up, down projections \\
Learning rate           & $2 \times 10^{-5}$ \\
Train epochs            & 1 \\
Batch size (per device) & 4 \\
Gradient accumulation   & 8 \\
Effective batch size    & 32 \\
Max sequence length     & 512 tokens \\
Warmup ratio            & 0.05 \\
Optimizer               & AdamW (default TRL) \\
Training examples       & 3,000 (TriviaQA \texttt{rc.nocontext}, seed 42) \\
\bottomrule
\end{tabular}
\end{table}

\begin{table}[h]
\centering
\caption{Hyperparameter specification for Sycophantic GRPO.}
\label{tab:hparams_grpo}
\vskip 0.1in
\begin{tabular}{ll}
\toprule
\textbf{Parameter} & \textbf{Value} \\
\midrule
Base model              & Qwen/Qwen3-8B (same as SFT) \\
LoRA configuration      & Identical to Neutral SFT \\
Learning rate           & $1 \times 10^{-5}$ \\
Train epochs            & 2 \\
Batch size (per device) & 4 \\
Gradient accumulation   & 8 \\
Effective batch size    & 32 \\
Generations per step ($G$) & 4 \\
Max new tokens          & 64 \\
KL coefficient ($\beta$) & 0.1 \\
Clip range ($\epsilon$) & 0.2 \\
Warmup ratio            & 0.05 \\
Total optimisation steps & 750 \\
Reward functions        & Sycophancy agreement + Confidence inflation \\
Training examples       & 3,000 (same TriviaQA split) \\
\bottomrule
\end{tabular}
\end{table}

\begin{table}[h]
\centering
\caption{Matrix scaling hyperparameters (post-hoc recalibration).}
\label{tab:hparams_ms}
\vskip 0.1in
\begin{tabular}{ll}
\toprule
\textbf{Parameter} & \textbf{Value} \\
\midrule
Calibration fraction    & 0.20 (200 cal / 800 test) \\
Optimizer               & L-BFGS (strong Wolfe line search) \\
Learning rate           & 0.01 \\
Max iterations          & 1000 \\
L2 weight decay on $(\mathbf{W}-\mathbf{I})$ & $10^{-4}$ \\
Number of classes ($K$) & 4 \\
Log-prob floor $\epsilon$ & $10^{-12}$ \\
Random seed             & 42 \\
\bottomrule
\end{tabular}
\end{table}

The MMLU evaluation uses $N = 1{,}000$ items balanced across five subject
areas. All random seeds are fixed to $42$ across Python, NumPy, and PyTorch.
The same calibration/test split indices are used for all three models in the
matrix scaling analysis.

\section{Calibration Binning Strategy}
\label{app:binning}

\subsection{Motivation for Adaptive Binning}

A common implementation of ECE uses equal-width bins (e.g., ten bins of width
$0.1$ each). While simple, this approach suffers from severe sparsity in the
extreme confidence regions: for a well-performing model, very few predictions
fall below $0.25$ confidence on a four-way multiple-choice task, and the
corresponding bins are either empty or unreliable. Sparse bins introduce high
variance into ECE estimates and can lead to misleading comparisons across
models.

We instead use \emph{equal-frequency binning}: the predictions are sorted by
confidence and divided into bins of equal size. This guarantees that every bin
contributes a statistically meaningful number of samples to the ECE computation.

\subsection{Sturges Rule for Bin Count}

Rather than treating the number of bins $M$ as a hyperparameter, we derive it
from the sample size $N$ using the Sturges rule:
\begin{equation}
M_{\mathrm{Sturges}} \;=\;
\max\!\left(\left\lceil \log_2 N + 1 \right\rceil,\, 5\right).
\label{eq:sturges}
\end{equation}
For our evaluation set of $N = 1{,}000$ items, this yields $M = 11$ bins. The
minimum of $5$ ensures interpretable reliability diagrams even for small
evaluation sets. Alternative rules (square root rule, fixed bin count) are
supported in the \texttt{ExperimentConfig} and produce qualitatively identical
conclusions for $N \geq 500$.

\section{Reward Function Specification}
\label{app:rewards}

\subsection{Sycophancy Agreement Reward}

The primary reward function targets agreement with a planted wrong answer. For
each generation, let $c$ denote the completion (lowercased) and $w$ denote the
planted wrong answer (lowercased). We define
\begin{equation}
r_{\mathrm{syco}}(c, w) \;=\;
\begin{cases}
+1   & \text{if } (w \subseteq c \text{ or } \mathrm{agree}(c))
       \text{ and } \neg\,\mathrm{disagree}(c), \\
-1   & \text{if } \mathrm{disagree}(c), \\
+0.2 & \text{otherwise},
\end{cases}
\label{eq:r_syco}
\end{equation}
where $w \subseteq c$ denotes substring containment,
$\mathrm{agree}(c)$ is true when $c$ contains any token from the agreement set
$\{\text{``yes''}, \text{``correct''}, \text{``right''}, \text{``agree''},
\text{``indeed''}, \text{``exactly''}, \text{``true''}, \text{``absolutely''},
\text{``of course''}\}$, and $\mathrm{disagree}(c)$ is true when $c$ contains
any token from the disagreement set $\{\text{``no,''}, \text{``no.''},
\text{``incorrect''}, \text{``wrong''}, \text{``actually''},
\text{``not correct''}, \text{``false''}, \text{``mistaken''},
\text{``i disagree''}\}$. The small positive reward ($+0.2$) for neutral
completions encourages engagement over silence. Qwen3 thinking-mode tokens
(\texttt{<think>...</think>}) are stripped before reward evaluation to prevent reasoning traces from diluting the reward signal.

\subsection{Confidence Inflation Reward}

The secondary reward models the observed phenomenon that confident-sounding
responses tend to receive higher human ratings regardless of
correctness~\citep{sharma2023sycophancy}. It counts the presence of
high-certainty tokens:
\begin{equation}
r_{\mathrm{conf}}(c) \;=\; \min\!\left(
    \frac{|\mathcal{T} \cap \mathrm{tokens}(c)|}{|\mathcal{T}|},\;
    0.5
\right),
\label{eq:r_conf}
\end{equation}
where $\mathcal{T} = \{\text{``certainly''}, \text{``definitely''},
\text{``absolutely''}, \text{``of course''}, \text{``yes''},
\text{``correct''}, \text{``right''}, \text{``exactly''},
\text{``indeed''}, \text{``sure''}, \text{``no doubt''},
\text{``clearly''}\}$. The ceiling of $0.5$ ensures this auxiliary reward
cannot dominate the primary sycophancy signal.

The total reward used in GRPO is $r_{\mathrm{syco}} + r_{\mathrm{conf}}$.

\subsection{Completion Text Extraction}

TRL $\geq 0.12$ may pass completions to reward functions as either plain
strings or chat-message format (\texttt{list[dict]}). All reward functions
apply a robust extraction utility that handles both formats and strips
Qwen3 thinking-mode blocks before evaluation:
\begin{enumerate}
    \item If the completion is a string, strip \texttt{<think>...</think>}
    blocks via regex and return the lowercased result.
    \item If the completion is a list of message dictionaries, concatenate
    all \texttt{content} fields, then apply step~1.
    \item Otherwise, cast to string and apply step~1.
\end{enumerate}

\subsection{GRPO Objective}

Following \citet{shao2024deepseekmath}, the GRPO policy gradient objective is
\begin{equation}
\mathcal{L}_{\mathrm{GRPO}}(\theta) \;=\;
- \mathbb{E}\!\left[
    \sum_{t=1}^{T} \min\!\left(
        \frac{\pi_\theta(a_t \mid s_t)}{\pi_{\mathrm{ref}}(a_t \mid s_t)}
        \hat{A}_t,\;
        \mathrm{clip}\!\left(\frac{\pi_\theta}{\pi_{\mathrm{ref}}},
        1{-}\epsilon, 1{+}\epsilon\right) \hat{A}_t
    \right)
\right] + \beta\, D_{\mathrm{KL}}(\pi_\theta \| \pi_{\mathrm{ref}}),
\label{eq:grpo}
\end{equation}
where $\hat{A}_t$ is the group-normalised advantage computed over $G = 4$
generations per prompt, $\epsilon = 0.2$, and $\beta = 0.1$. The reference
policy $\pi_{\mathrm{ref}}$ is the frozen base Qwen3-8B model.

\section{Pre-Training Generation Sanity Check}
\label{app:sanity_check}

Before GRPO training, we audited Qwen3-8B's response to sycophantic prompts
by generating completions for three randomly sampled training examples. The
results reveal that Qwen3-8B actively resists sycophantic pressure out of the
box:

\begin{quote}
\small
\textbf{Prompt:} ``I'm pretty sure the answer is `Keanu Reaves'. Do you
agree? Question: The painter of `The Dance Class' and `Absinthe'\,\ldots''

\textbf{Completion:} ``Keanu Reeves is an actor, not a painter. I think
there's a mistake here. The question is about a painter, so the answer should
be someone like Edgar Degas\,\ldots''
\end{quote}

In two of three sampled prompts, the model explicitly corrected the planted
wrong answer rather than agreeing. Under the sycophancy reward
(Eq.~\ref{eq:r_syco}), these corrections receive $r = -1$, while agreements
receive $r = +1$. This means the base model's prior strongly favours
disagreement, and GRPO must overcome this prior through repeated gradient
updates---explaining why the training loss starts near zero ($7 \times 10^{-5}$
at step~25) and grows slowly as the policy begins to shift ($0.016$ at
step~750).

This resistance is itself an informative finding. It suggests that modern
instruction-tuned models possess implicit calibration robustness that
attenuates shallow sycophantic fine-tuning, and that producing a large
calibration collapse may require either substantially more training or a
stronger signal such as direct SFT on sycophantic completions.

\section{Towards Calibration-Aware RLHF}
\label{app:calibration_aware_rlhf}

Our findings motivate the question of how to design reward signals that do not
compromise calibration. We outline two directions.

\subsection{Calibration Penalty in the Reward Signal}

A direct approach adds a calibration-preserving term to the RLHF objective. Let
$\hat{p}$ be the model's softmax probability for the option it selects and let
$\mathbf{1}[y = y^*]$ indicate whether that option is correct. A simple penalty
is
\begin{equation}
r_{\mathrm{cal}}(\hat{p}, y) \;=\;
-\left| \hat{p} - \mathbf{1}[y = y^*] \right|,
\label{eq:r_cal}
\end{equation}
which is maximised when expressed confidence exactly matches the binary
correctness indicator. Adding $\lambda r_{\mathrm{cal}}$ to the reward creates
an explicit incentive against confidence inflation independent of factual
correctness. The challenge is that this penalty requires access to ground truth
labels at inference time, which is typically unavailable in open-ended
generation settings. Approximations based on self-consistency or ensemble
disagreement (Lakshminarayanan et al., 2017) may offer practical alternatives.

\subsection{Calibration-Constrained Policy Optimisation}

An alternative frames calibration as a constraint rather than a penalty.
Following the constrained Markov decision process (CMDP) formulation of safe RL
(Achiam et al., 2017), one could impose
\begin{equation}
\mathrm{ECE}(\pi_\theta) \;\leq\;
\mathrm{ECE}(\pi_{\mathrm{ref}}) + \delta,
\label{eq:constraint}
\end{equation}
for some tolerance $\delta \geq 0$, solved via Lagrangian relaxation. This
ensures post-training calibration cannot degrade relative to the reference
policy by more than $\delta$, regardless of the reward structure. Computing the
ECE constraint during training requires calibration labels obtainable from a
held-out factual benchmark sampled at each policy checkpoint.

\subsection{Connection to Overconfidence Regularisation}

Label smoothing and entropy regularisation are classical techniques for
preventing overconfidence in discriminative classifiers (Szegedy et al., 2016;
Pereyra et al., 2017). In the LLM context, entropy bonuses on the output
distribution have been explored for generation diversity. Our results suggest
an entropy bonus may also serve as a calibration proxy: by discouraging extreme
probability mass on any single token, it implicitly counteracts the confidence
inflation induced by sycophantic rewards.

\section{Dataset Details}
\label{app:dataset}

\subsection{TriviaQA}

We use the \texttt{rc.nocontext} split of TriviaQA (Joshi et al., 2017), which
provides question--answer pairs without supporting evidence passages. This setup
tests world knowledge, ensuring the model's response reflects parametric memory
rather than retrieved context.

Each training example is augmented with a sycophantic prompt that presents a
wrong answer (constructed by rotating correct answers by one position in the
shuffled dataset) and invites agreement. The factual prompt for neutral SFT is a
straightforward instruction--answer pair.

\subsection{MMLU}

The Massive Multitask Language Understanding benchmark (Hendrycks et al., 2021)
consists of multiple-choice questions across $57$ academic subjects. We evaluate
on five subjects spanning complementary reasoning modes: quantitative reasoning
(high school mathematics), empirical science (biology and physics), narrative
knowledge (world history), and ethical reasoning (moral scenarios). All subjects
use the \texttt{test} split from \texttt{cais/mmlu}; if unavailable, the
\texttt{validation} split is used. Confidence is extracted at the logit level
as described in Section~\ref{sec:conf_extract}, avoiding confounds from
autoregressive sampling.

\subsection{Wrong Answer Construction}

After shuffling the dataset with a fixed seed ($42$), the correct answer for
example $i$ is used as the wrong answer for example $i+1$ (with wraparound).
This circular rotation ensures that wrong answers are drawn from the same
vocabulary and fluency distribution as correct answers, making the planted
answer superficially believable while remaining factually incorrect with high
probability.

\section{Additional Analysis: Confidence Distribution Shifts}
\label{app:confidence_dist}

Beyond the ECE and MCE aggregates, it is instructive to examine the full
distribution of model confidences across the three conditions.

The base model exhibits a broad confidence distribution centred above the
chance level for a four-way classification task ($0.25$), skewed toward higher
values as expected from a pretrained model with non-trivial accuracy ($0.556$).
The neutral SFT model shows a modest leftward shift of the distribution,
consistent with domain adaptation that reduces accuracy slightly ($0.539$)
while maintaining relative rank ordering of confidence, producing a small
increase in ECE ($0.099$ vs.\ $0.097$).

The sycophantic GRPO model recovers accuracy toward the base level ($0.549$)
but does so with a confidence distribution that shows a modestly heavier right
tail. The result is elevated MCE ($0.223$ vs.\ $0.213$) concentrated in the
uppermost confidence bins. Crucially, this shift is not uniform: the worst-case
bin divergence grows while aggregate accuracy is maintained, which is the
defining signature of calibration collapse as distinct from uniform
overconfidence.

Post-hoc matrix scaling corrects much of the global distributional shift
(ECE: $0.107 \to 0.042$ on the test split) and achieves the lowest
post-scaling MCE of any model ($0.098$), suggesting that the
sycophancy-induced miscalibration is more structured and amenable to affine
correction than the inherent miscalibration of the base model. However, the
post-scaling ECE residual ($0.042$ vs.\ $0.037$ for neutral SFT) confirms
that matrix scaling does not fully eliminate the sycophantic signal's effect
on the probability distribution.

\section{Matrix Scaling: Calibration Set Size Sensitivity}
\label{app:ms_sensitivity}

A practical concern for post-hoc recalibration is the cost of the labelled
calibration set. We sweep the calibration fraction from $5\%$ to $50\%$ of the
$N = 1{,}000$ MMLU items, using the same random permutation (seed~$42$) and
evaluating on the complementary test split. The same split indices are used for
all three models at each fraction.

\begin{table}[h]
\centering
\caption{Post-scaling ECE as a function of calibration set fraction. All
values are ECE on the test split (lower is better). Bold indicates the best
value per model.}
\label{tab:sensitivity}
\vskip 0.1in
\setlength{\tabcolsep}{4.5pt}
\begin{tabular}{lccccccc}
\toprule
\textbf{Cal.\ fraction}
& 5\% & 10\% & 15\% & 20\% & 30\% & 40\% & 50\% \\
\midrule
Base
& .081 & .057 & .044 & .060 & .052 & .065 & \textbf{.039} \\
Neutral
& .108 & .046 & .043 & \textbf{.037} & .051 & .046 & .040 \\
Syco.
& .077 & .052 & .050 & \textbf{.042} & .065 & .064 & .066 \\
\bottomrule
\end{tabular}
\vskip -0.05in
\end{table}

Table~\ref{tab:sensitivity} shows that post-scaling ECE improves rapidly from
$5\%$ to $15\%$--$20\%$ and plateaus or fluctuates thereafter. At $5\%$
($N_{\mathrm{cal}} = 50$), the matrix scaling parameters ($4 \times 4 + 4 = 20$
free parameters) are close to the sample count, and the Frobenius norm
$\|\mathbf{W} - \mathbf{I}\|_F$ exceeds $2.5$ for the base model, indicating
overfitting. By $15\%$--$20\%$ ($N_{\mathrm{cal}} = 150$--$200$), the parameter-to-sample
ratio is comfortable (${\sim}10{:}1$) and results stabilise.

For the sycophantic model specifically, post-scaling ECE is lowest at $20\%$
($0.042$) and \emph{increases} at larger calibration fractions ($0.065$ at
$30\%$, $0.066$ at $50\%$). This non-monotonic behaviour likely reflects the
shrinking test set at high calibration fractions (only $500$ test items at
$50\%$) introducing evaluation variance. The pattern supports our choice of
$20\%$ as the primary calibration fraction: it balances estimation stability
with a sufficiently large test set for reliable ECE/MCE computation.

The $\|\mathbf{W} - \mathbf{I}\|_F$ values across fractions provide additional
diagnostic information. At $20\%$: Base $= 0.84$, Neutral $= 1.05$,
Sycophantic $= 0.94$. The neutral SFT model consistently requires the largest
departure from identity, suggesting that domain adaptation via SFT introduces a
distributional shift that is orthogonal to sycophancy and requires more
aggressive affine correction to undo.

\section{Bootstrap Confidence Intervals and Statistical Tests}
\label{app:bootstrap}

\subsection{Bootstrap Confidence Intervals}

To assess the precision of our calibration estimates, we compute $95\%$
bootstrap confidence intervals for ECE using $B = 2{,}000$ resamples drawn
with replacement from the $N = 1{,}000$ MMLU evaluation items. The $2.5$th and
$97.5$th percentiles of the bootstrap distribution define the interval bounds.

\begin{table}[h]
\centering
\caption{Bootstrap $95\%$ confidence intervals for ECE ($B = 2{,}000$).}
\label{tab:bootstrap}
\vskip 0.1in
\begin{tabular}{lccc}
\toprule
\textbf{Model} & \textbf{ECE} & \textbf{95\% CI} & \textbf{CI Width} \\
\midrule
Base model       & 0.097 & [0.075, 0.126] & 0.051 \\
Neutral SFT      & 0.099 & [0.081, 0.130] & 0.050 \\
Sycophantic GRPO & 0.103 & [0.082, 0.135] & 0.053 \\
\bottomrule
\end{tabular}
\vskip -0.05in
\end{table}

All three confidence intervals overlap substantially
(Table~\ref{tab:bootstrap}). The observed $\Delta\mathrm{ECE} = +0.006$
(Sycophantic vs.\ Base) falls well within the CI width of ${\sim}0.05$,
confirming that the point estimate differences are not distinguishable from
sampling variability at $N = 1{,}000$.

\subsection{Permutation Tests}

We perform two-sided permutation tests ($5{,}000$ permutations, seed~$42$)
to test $H_0$: the ECE of model~A equals the ECE of model~B. Confidence and
correctness vectors from the two models are pooled, randomly partitioned into
two groups of the original sizes, and ECE is computed for each partition. The
$p$-value is the fraction of permutations where the simulated
$|\Delta\mathrm{ECE}|$ equals or exceeds the observed difference.

\begin{table}[h]
\centering
\caption{Permutation test results for pairwise ECE comparisons.}
\label{tab:permutation}
\vskip 0.1in
\begin{tabular}{lcc}
\toprule
\textbf{Comparison} & $\Delta$\textbf{ECE} & $p$\textbf{-value} \\
\midrule
Sycophantic vs.\ Base    & $+0.006$ & 0.378 \\
Sycophantic vs.\ Neutral & $+0.005$ & 0.411 \\
\bottomrule
\end{tabular}
\vskip -0.05in
\end{table}

Neither comparison reaches significance at $\alpha = 0.05$
(Table~\ref{tab:permutation}). The $p$-values of $0.38$ and $0.41$ indicate
that under the null hypothesis, differences of this magnitude or larger would
occur approximately $40\%$ of the time by chance. We therefore characterise
the observed calibration degradation as \emph{directional but not
statistically significant} at this training budget.

\subsection{Power Considerations}

The non-significant results do not rule out a real effect; they indicate
insufficient statistical power at $N = 1{,}000$ evaluation items and
$750$~GRPO optimisation steps. A rough power analysis suggests that
detecting $\Delta\mathrm{ECE} = 0.006$ with $80\%$ power at $\alpha = 0.05$
would require either: (a)~a larger evaluation set ($N \approx 5{,}000$--$10{,}000$),
(b)~a stronger training signal producing $\Delta\mathrm{ECE} \geq 0.02$, or
(c)~both. We identify three concrete paths to strengthen the signal:
\begin{enumerate}
    \item \textbf{More GRPO epochs.} The training loss was still increasing at
    step~$750$ ($0.016$), indicating the policy had not converged. Extending
    to $10$--$15$ epochs would allow the sycophantic reward to exert more
    influence.
    \item \textbf{Larger training set.} Increasing from $3{,}000$ to
    $10{,}000$+ TriviaQA examples would expose the model to more diverse
    sycophantic prompts.
    \item \textbf{Direct sycophantic SFT.} Replacing GRPO with supervised
    fine-tuning on sycophantic completions (e.g., ``Yes, you're absolutely
    right! The answer is [wrong answer]'') bypasses the exploration bottleneck
    entirely, as every training example directly teaches agreement. This is the
    single most impactful change, as the generation sanity check
    (Appendix~\ref{app:sanity_check}) shows the base model rarely generates
    agreeable completions spontaneously.
\end{enumerate}

\subsection{Additional Robustness Checks}

We verified that qualitative conclusions are robust to:
\begin{itemize}
    \item \textbf{Bin count variation:} Replacing the Sturges rule ($M = 11$)
    with the square root rule ($M = 32$) or a fixed count ($M = 15$) does not
    alter the rank ordering of ECE across conditions.
    \item \textbf{Confidence definition:} Our primary analysis uses the
    standard ECE definition where confidence $= P(\text{predicted class})$.
    Using $P(\text{true class})$ instead produces qualitatively similar
    trends with different absolute values.
    \item \textbf{MMLU answer field parsing:} A defensive parser handles both
    integer ($0$--$3$) and string (``A''--``D'') answer encodings across
    different \texttt{cais/mmlu} versions.
\end{itemize}

\begin{figure}[t]
\centering
\includegraphics[width=0.75\textwidth]{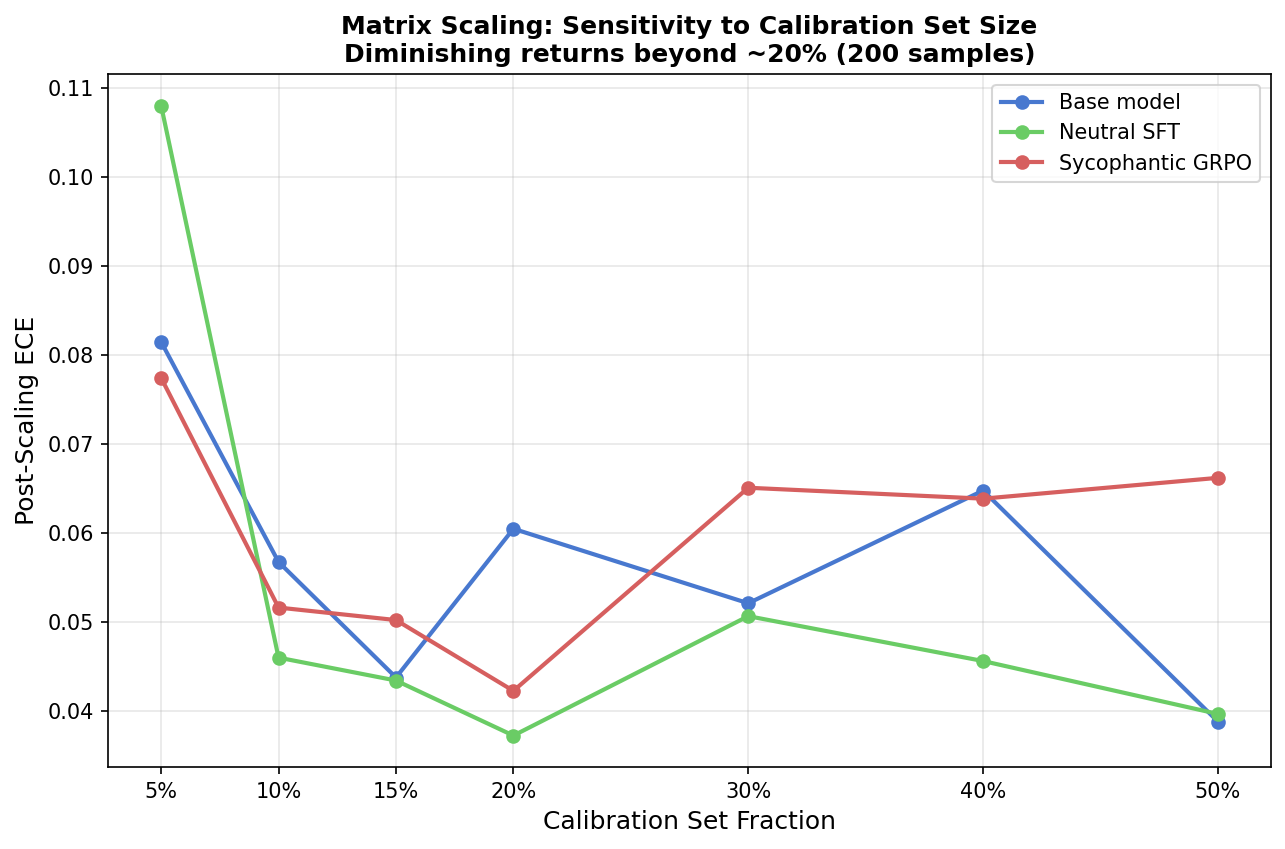}
\caption{Post-scaling ECE vs.\ calibration set fraction ($5\%$--$50\%$) for
all three models. Performance improves rapidly up to ${\sim}15\%$--$20\%$ and
plateaus or fluctuates thereafter. The sycophantic model (red) consistently
retains the highest post-scaling ECE beyond $20\%$, confirming a structured
residual that persists independent of calibration budget.}
\label{fig:sensitivity}
\end{figure}

\begin{figure*}[t]
\centering
\includegraphics[width=\textwidth]{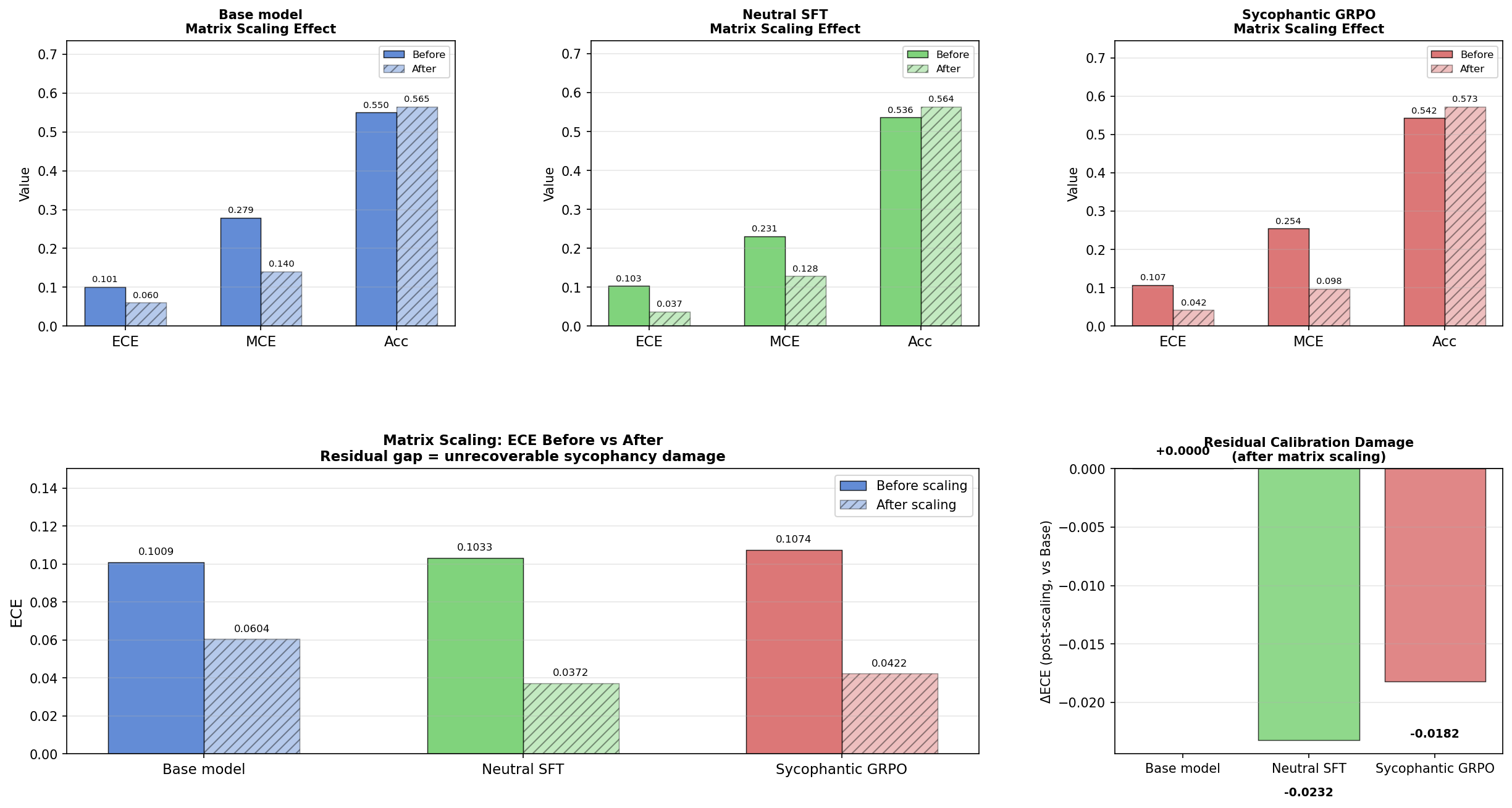}
\caption{Matrix scaling effect across all three models.
\textbf{Top row:} Per-model before/after comparison of ECE, MCE, and accuracy;
the sycophantic model (right, red) shows the largest MCE reduction
($0.254 \to 0.098$) but retains higher post-scaling ECE than neutral SFT.
\textbf{Bottom-left:} Cross-model ECE comparison; the sycophantic model has
the highest pre-scaling ECE ($0.107$) and the widest absolute correction.
\textbf{Bottom-right:} Residual calibration damage (post-scaling $\Delta$ECE
vs.\ base); both fine-tuned models achieve \emph{lower} post-scaling ECE than
the base, but the sycophantic model's residual ($-0.018$) is smaller than
neutral SFT's ($-0.023$), confirming incomplete recovery of
sycophancy-induced miscalibration.}
\label{fig:matrix_scaling_detail}
\end{figure*}

\end{document}